# How Useful are LLMs for Grammar Engineering? Cantonese ParGram Resources and Controlled Experimental Evaluation with English Baselines

**Chit-Fung Lam**
University of Oxford
University of Manchester
`lawrence.lam@ling-phil.ox.ac.uk`

## Abstract

This paper presents new Cantonese ParGram resources and evaluates LLMs for knowledge-driven grammar engineering within a controlled experimental paradigm. Using Cantonese ParGram resources as gold standards, with corresponding English baselines, we investigate whether OpenAI's `gpt-oss-120b` and `GPT-5.4` can generate machine-processable grammars from sentences and target formal structures under systematically varied prompting conditions. `GPT-5.4` outperformed `gpt-oss-120b`, while grammars generated from target formal structures generally outperformed those generated from sentences. Although both models could generate locally plausible phrase-structure rules, lexical entries, and templates, they often struggled to coordinate interacting formal constraints, especially in multi-construction settings. The results characterize both the capabilities and limitations of current LLMs for potential integration into AI-assisted expert workflows: LLMs may support intermediate stages of grammar development, but human linguistic expertise remains central to analysis, validation, and refinement. The study also contributes new Cantonese symbolic grammatical resources.

## 1 Introduction

Grammar engineering involves the design and computational implementation of formal grammars (Bender, 2008; Duchier and Parmentier, 2015). Within the Parallel Grammar (ParGram) Project (Butt et al., 1999, 2002; Sulger et al., 2013), broad-coverage grammars are developed in parallel across languages using the Lexical–Functional Grammar (LFG) formalism (Bresnan et al., 2016; Dalrymple et al., 2019) as computationally implemented in Xerox Linguistic Environment (XLE) (Crouch et al., 2011). These grammars then generate parsebanks containing c(onstituent)-structures and f(unctional)-structures (Section 2.2), collectively forming the ParGramBank system. ParGram treebanks are pure parsebanks in that they are generated automatically by parsing sentences using their respective LFG/XLE grammars, without ad hoc corrections or manual annotation (Rosén, 2023).

The purpose of this paper is twofold:

1. It presents new Cantonese treebank resources as part of the ParGram Project;
2. It presents a controlled experimental evaluation of the capabilities and limitations of current LLMs in supporting knowledge-driven grammar engineering, using Cantonese ParGram resources as gold standards with corresponding English baseline conditions.

For the second objective, we evaluate the extent to which large language models (LLMs) can create XLE-style phrase structure rules, functional annotations, templates, and lexical entries required for XLE grammars (Crouch et al., 2011). Our experiments are motivated by recent interest in grammar creation using LLMs (Spencer and Kongborrirak, 2025) and the potential incorporation of LLMs into traditional grammar engineering workflows (Lam and Uí Dhonnchadha, 2026).

Knowledge-driven grammar engineering typically requires substantial manual effort as well as expertise in both linguistic formalism and computational implementation (Duchier and Parmentier, 2015). LLMs can potentially reduce the grammar engineering costs by assisting with rule generation, lexical specification, and structural generalization across languages. However, it remains unclear how useful current LLMs are in generating grammars that are linguistically adequate, formally accurate, and computationally usable within existing symbolic grammar engineering pipelines (Lam and Uí Dhonnchadha, 2026). To address these questions, we conducted experiments using Ope-nAI's gpt-oss-120b and GPT-5.4 with prompting strategies based on one-shot in-context learning (Brown et al., 2020). Our evaluation focused on the structural validity, linguistic adequacy, and formal accuracy of the generated grammars.

The study adopts a controlled experimental paradigm rather than attempting to reproduce end-to-end development of a mature ParGram grammar. We systematically vary the source of grammatical information, the scope of grammar generation, and the language model while holding the remaining aspects of the experimental setting constant. Here, 'usefulness' is operationalized in terms of rule-level formal accuracy, parse-level structural quality, and error patterns under these controlled conditions, rather than end-to-end development efficiency. This design allows us to characterize where current LLMs succeed and fail in symbolic grammar generation and what these patterns imply for AI-assisted, expert-led grammar-engineering workflows.

## 2 Background and Related Work

### 2.1 Grammar engineering and ParGram

Grammar engineering is a practice used by a number of linguistic communities, including LFG and Head-driven Phrase Structure Grammar (HPSG; Pollard and Sag 1994), where formal grammars are computationally implemented for linguistic analysis and NLP applications (Forst and King, 2023; Zamaraeva et al., 2022; Bender and Emerson, 2021; Müller, 2015). From a theoretical perspective, implemented grammars help test linguistic hypotheses and capture complex interactions among linguistic constraints (Bender, 2008; Forst and King, 2023). In parallel grammar projects such as ParGram, parsed treebanks provide insights into cross-linguistic similarities and differences (Butt et al., 1999). From an applied perspective, implemented grammars have been used in applications including grammar checkers, semantic engines, dialogue systems, computer-assisted language learning, and, more recently, the generation of well-annotated training data for machine-learning NLU models (Forst and King, 2023).

ParGramBank was initially developed as a collection of parallel LFG treebanks covering twelve languages from six language families, based on a shared test suite of sentences (Butt et al., 1999). The first 50 sentences (Appendix A), which form the focus of this paper, include a broad range of grammatical phenomena such as transitivity, unaccusative and unergative predicates, clause types, voice alternations, dative and double-object constructions, complementation, control and raising, relative clauses, negation, phrasal verbs, prepositional phrases, copular constructions, reflexives, reciprocals, modification, expletives, and existentials (Sulger et al., 2013). These sentences are translated across the different ParGram languages to form a parallel corpus through which grammar engineers incrementally implement and extend grammatical coverage for the wide range of syntactic constructions. Consistent with the approaches advocated by Lehmann et al. (1996) and Bender et al. (2011), ParGram places strong emphasis on achieving broad grammatical coverage in grammar engineering (Sulger et al., 2013).

### 2.2 LFG formalism and XLE

LFG is a constraint-based grammatical formalism with a modular architecture in which different linguistic representations are related through mathematically well-defined functions (Dalrymple and Findlay, 2019).[1] Its syntactic component consists of two levels of representation: c-structure and f-structure. C-structures encode constituency, word order, and part-of-speech information, whereas f-structures represent grammatical functions such as SUBJ and OBJ together with associated morphosyntactic features (Bresnan et al., 2016; Börjars, 2020). Appendix B illustrates the mapping algorithm between c- and f-structures, showing how an f-structure is formed through the satisfaction of functional equations projected from lexical entries and functional annotations associated with c-structure rules (Bresnan et al., 2016). XLE is a grammar engineering environment for implementing LFG-based grammars. It contains parsing and generation algorithms for LFG grammars together with graphical interfaces for grammar writing, testing, and debugging (Crouch et al., 2011). All the prompts used in our study include an example English toy grammar in XLE format (Appendices C–F). Figure 2 in Section 3.1 provides a concrete Cantonese example in which the corresponding c- and f-structures are shown together.

[1]For accessible introductions to LFG, see Börjars (2020) and Dalrymple and Findlay (2019); more comprehensive materials are provided by Bresnan et al. (2016) and Dalrymple et al. (2019).

### 2.3 LFG grammar engineering and LLMs

Spencer and Kongborrirak (2025) investigated the creation of LFG/XLE grammars for the low-resource language Moklen using prompt-based in-context learning with OpenAI's `gpt-4o-mini`. Although grammar-rule accuracy formed part of the study, the published evaluation focused primarily on Moklen–English translation performance. Quantitative evaluation of grammar-rule accuracy was limited, with an 86/100 accuracy score reported for lexical entries. The study did not systematically evaluate the formal well-formedness or computational parsability of the generated grammars, nor the quality of the syntactic structures produced when parsing with the LLM-generated grammars. Nevertheless, the work represents a very important early exploration of the potential use of LLMs for grammar engineering in under-resourced languages.

Lam and Uí Dhonnchadha (2026) evaluated the ability of OpenAI's `gpt-oss-120b` to generate LFG f-structures from Cantonese and Irish ParGram sentences in support of cross-linguistic grammar engineering. The study proposed a detailed four-point assessment scale (Appendix G) for evaluating the quality of the generated f-structures. The results showed that the model could produce around 70% of outputs capturing key predicate–argument relations and that LLMs may suggest good alternative analyses. However, the study did not directly evaluate the model's ability to generate XLE grammars, including the formal constraints and rule interactions in grammar development, focusing instead on the model's ability to generate f-structures from Cantonese and Irish sentences provided in the prompts.

In Section 4, we address these limitations through a controlled experimental evaluation of LLM capabilities in generating Cantonese XLE grammars, including phrase-structure rules, functional annotations, templates, and lexical entries. Besides assessing formal and notational accuracy, we also evaluate the structural consistency and linguistic adequacy of the LLM-generated XLE grammars. Our study emphasizes grammar-oriented evaluation using symbolic Cantonese ParGram resources as gold standards, alongside English baseline conditions.

## 3 New Cantonese ParGram Resources

Despite being widely spoken, Cantonese remains under-resourced in computational linguistics (Xiang et al., 2024), with relatively limited formal grammatical resources available. This section presents new Cantonese ParGram resources, focusing on the treebank. Similar to other ParGram grammars, the Cantonese ParGram grammar is a handcrafted grammar implemented in XLE. The treebank is then automatically generated by parsing ParGramBank sentences using the Cantonese ParGram grammar (Section 2.1; Appendix A).

This section describes selected grammatical phenomena represented in the treebank structures among the first 50 ParGram sentences. Where appropriate, we discuss the XLE rules responsible for the successful parsing of the relevant constructions. Our discussion focuses on f-structures, which represent the deeper syntactic relations of the sentences (Section 2.2), consistent with ParGram practice where grammar engineers place emphasis on cross-linguistic comparisons at the f-structure level (Sulger et al., 2013). These treebank structures, together with their Cantonese ParGram XLE rules, serve as important gold standards for evaluating LLM outputs in Section 4.

### 3.1 Selected phenomena: theoretical insights and computational implementation

A guiding principle in ParGram (and LFG) is that, while c-structures may vary substantially across languages, f-structures maintain a good degree of cross-linguistic parallelism, particularly in the encoding of core grammatical functions (GFs) such as SUBJ and OBJ, alongside language-specific properties. This principle has been demonstrated in numerous ParGram studies (e.g., Sulger et al., 2013; Lam and Uí Dhonnchadha, 2026). Another important principle, which is our focus here, concerns the incorporation of insights from theoretical linguistics while maintaining computational robustness in implementation. In the following discussion, we present several cases illustrating how the second principle is realized in Cantonese ParGram with respect to a few linguistic phenomena involving certain language-specific properties: (i) control construction; (ii) object-fronting construction; and (iii) passive construction.

**Control construction** In LFG, there are two main types of control mechanisms: functional control and anaphoric control (Bresnan, 1982; Bres-nan et al., 2016; Dalrymple et al., 2019). In the f-structure, functional control involves structure sharing between the embedded SUBJ controllee and a matrix controller function, whereas in anaphoric control, the controllee is represented as a pronoun <PRED, 'PRO'>. Functional control involves an open complement function XCOMP, while anaphoric control involves a closed complement function COMP. Determining whether a control verb licenses functional or anaphoric control often requires language-specific evidence. In Cantonese, recent theoretical studies (Lam, 2026; Lam and Dalrymple, 2026) suggest that many control verbs, such as *try* and *prepare*, involve anaphoric control, whereas transitive subject-control verbs such as *promise* involve functional control. This theoretical distinction is encoded in the Cantonese ParGram grammar and treebank. As an illustration, Figure 1 shows the f-structure of the ParGram sentence in (1), containing the functional control verb 應承 'promise', where the matrix SUBJ shares the same structure as the embedded SUBJ.[2] The XLE implementation involves a structure-sharing equation: `(^SUBJ)=(^XCOMP SUBJ)` (Crouch et al., 2011).

(1) 個 農夫-嘅 女 應承 修理
CL farmer-POSS daughter promise repair
架 拖拉機
CL tractor
'The farmer's daughter promised to repair the tractor.' (ParGramBank sentence 49)

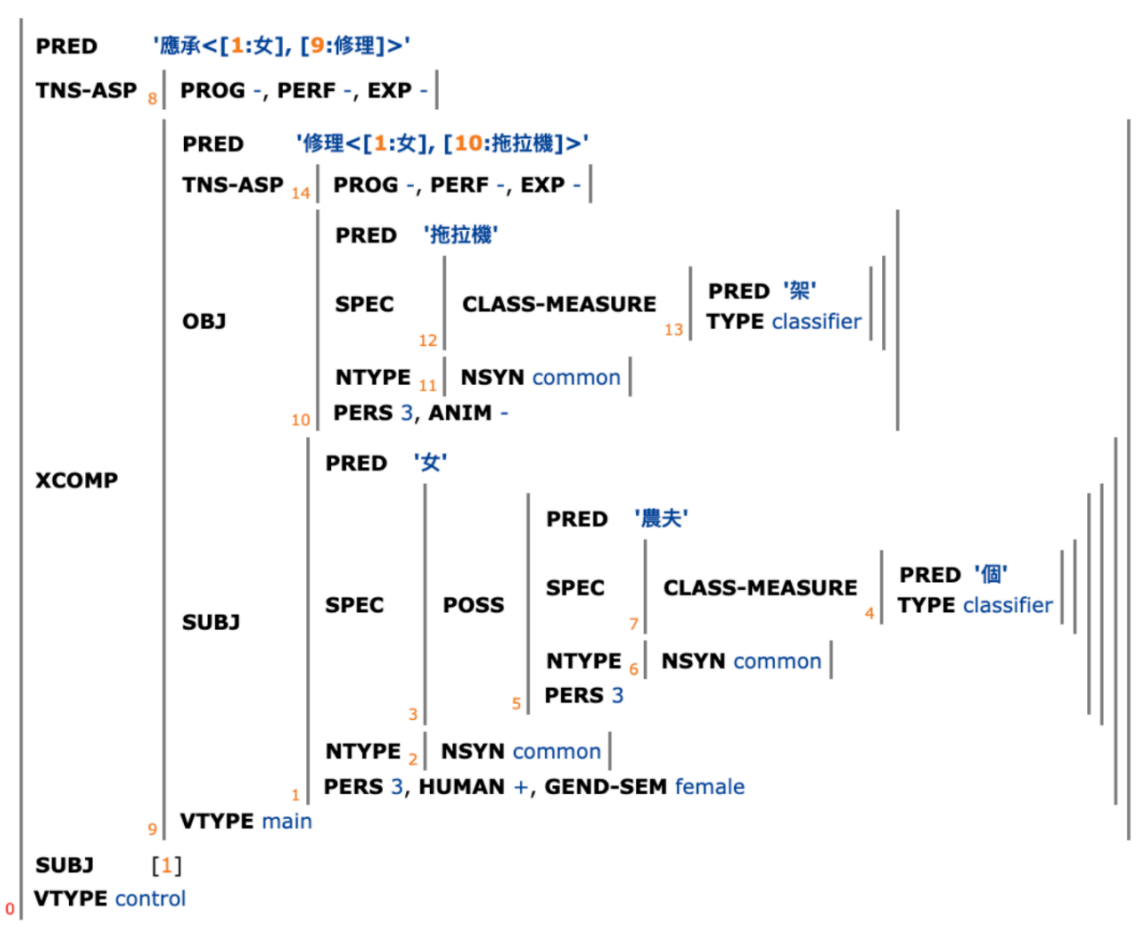

Figure 1: F-structure of example (1).

[2]The f-structures in this paper were generated using the online INESS interface (Rosén et al., 2012), following the upload of the Cantonese XLE grammar to the server with assistance from Paul Meurer.

**Object-fronting construction** Cantonese is an SVO language. The Cantonese 將 *zoeng* is equivalent to the Mandarin Chinese 把, functioning as an OBJ-fronting marker that preposes the OBJ to a preverbal position (Lam et al., 2023; Huang et al., 2009). Example (2) illustrates the alternation between a canonical SVO structure and an OBJ-fronting structure. Figure 2 shows the c- and f-structures of (2b).

(2) a. 個 農夫 斬 棵 樹 落嚟
CL farmer cut CL tree down
'The farmer cut the tree down.'

b. 個 農夫 將 棵 樹 斬 落嚟
CL farmer ZOENG CL tree cut down
'The farmer cut the tree down.' (ParGramBank sentence 8)

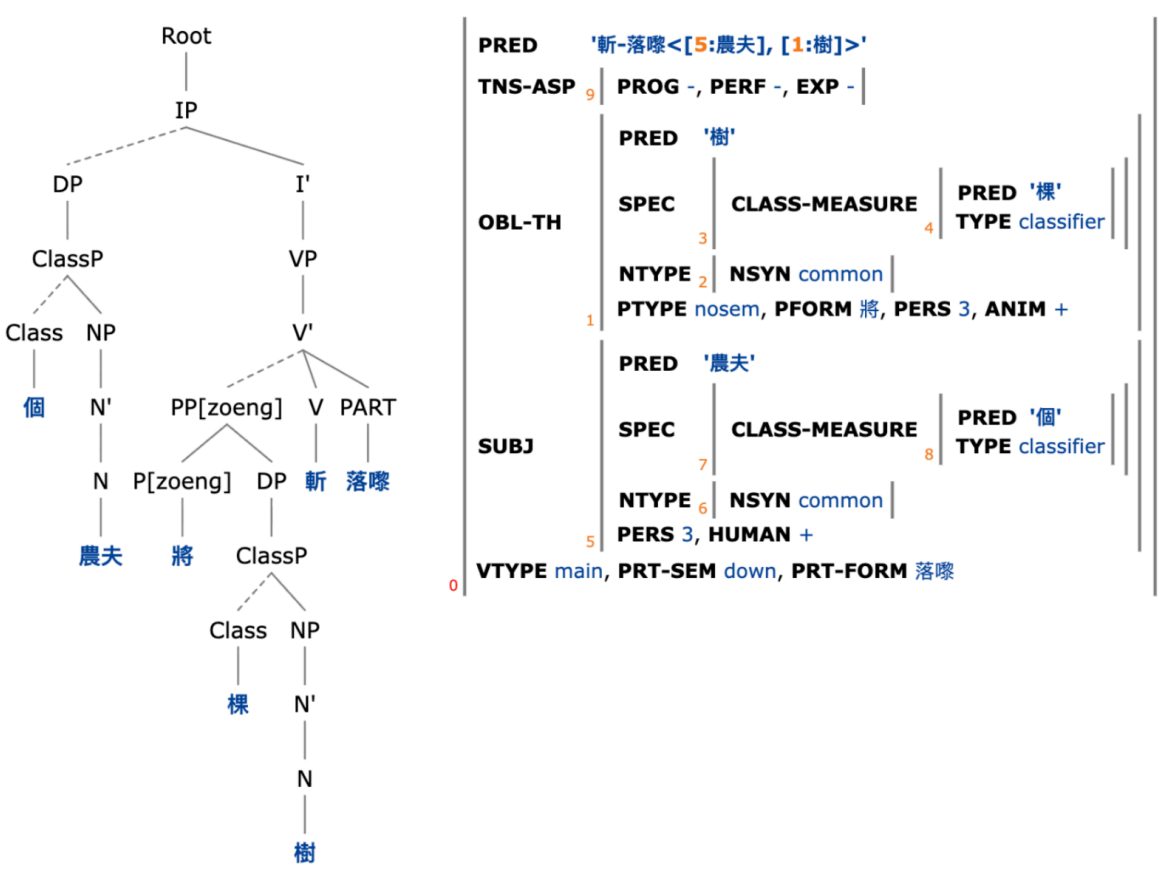

Figure 2: C- and f-structures of example (2b).

To account for the alternation between the two structures, the Cantonese ParGram grammar adopts a lexical rule for GF alternation applied to the subcategorization frames of transitive verbs:

```
OBJ-FRONT(SCHEMATA)=
{SCHEMATA|SCHEMATA (^OBJ)-->(^OBL-TH)}.
```

Lexical rules are commonly used in cross-linguistic ParGram grammars to model GF alternations (Crouch et al., 2011), such as the English dative–ditransitive alternation. This rule, together with other rules in Cantonese ParGram, including c-structure rules for $PP_{[zoeng]}$, enables the grammar to parse both (2a) and (2b), leveraging the alternation relation between the two constructions.

**Passive construction** The Cantonese passive construction contains the passive marker 俾, an equivalent of Mandarin 被. Unlike Mandarin

被, however, Cantonese 俾 obligatorily requires an overt agent following the passive marker (Matthews and Yip, 2013). We extend the theoretical LFG analysis of 被 proposed by Her (2008), Lam et al. (2023), and Her (2009) to Cantonese 俾, analyzing it as a higher verb involving subordination rather than a preposition. Under this analysis, 俾 functions as a three-place predicate selecting for SUBJ, OBJ, and XCOMP. However, unlike Lam et al. (2023) and Her (2009), we simplify the analysis by omitting an embedded TOPIC function mediating between the matrix SUBJ and the embedded OBJ. Our analysis closely resembles that of Her (2008, p. 209), requiring structure sharing between the matrix SUBJ and the embedded OBJ, as well as between the matrix OBJ and the embedded SUBJ. This requires the structure-sharing equations `(^SUBJ)=(^XCOMP OBJ)` and `(^OBJ)=(^XCOMP SUBJ)`. Figure 3 shows the XLE implementation of this analysis for (3).

(3) 棵 樹 琴日 俾 人 斬-咗 落嚟
CL tree yesterday PASS person cut-PFV down
'The tree was cut down yesterday.' (ParGramBank sentence 11)

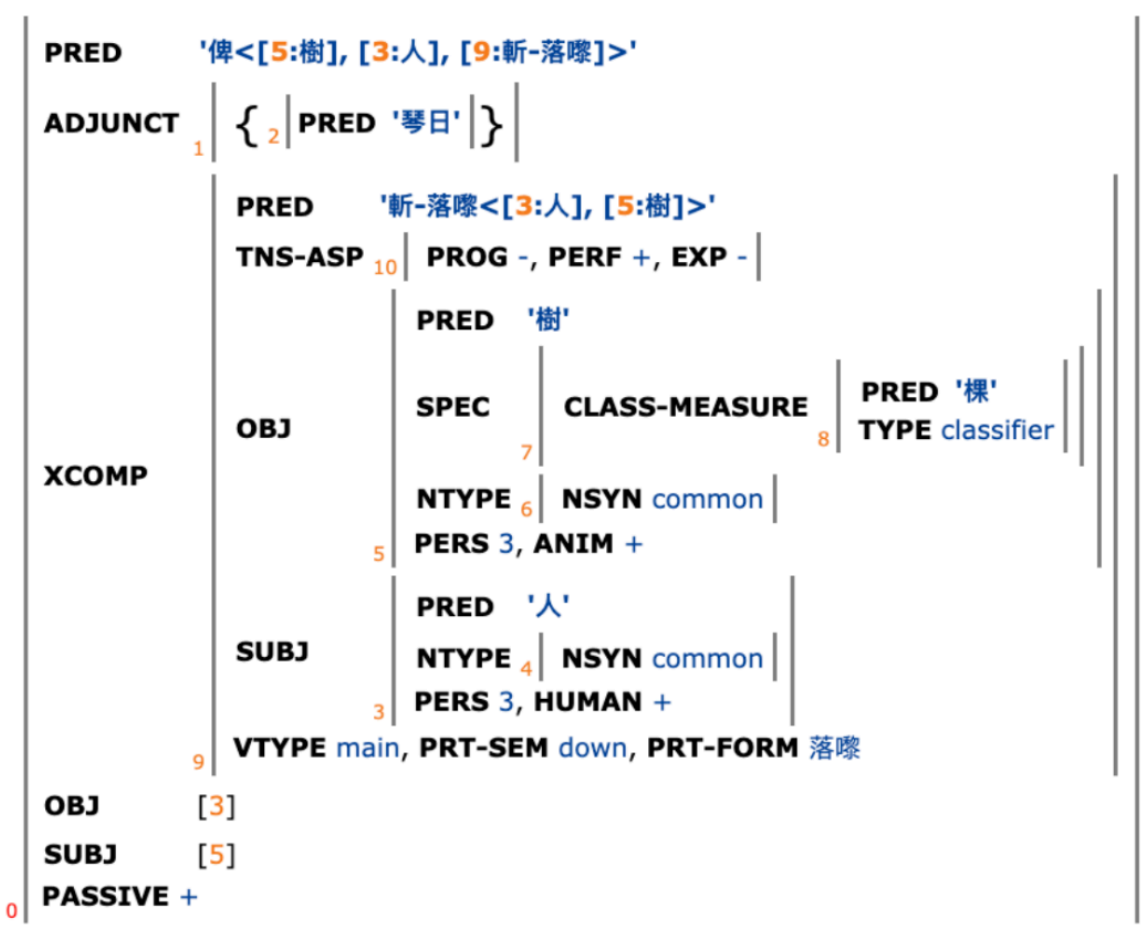

Figure 3: F-structure of example (3).

## 4 New Case Study: Grammar Engineering and LLMs

Using the newly developed Cantonese ParGram resources (Section 3) as gold standards, with corresponding English ParGram baselines, we conducted two controlled experiments to evaluate the capabilities and limitations of LLMs in generating XLE-style grammar components.

### 4.1 Methodology

**Model selection** We selected two OpenAI LLMs: `gpt-oss-120b` and `GPT-5.4`. `gpt-oss-120b` is one of only two open-weight models released by OpenAI (OpenAI et al., 2025) and offers greater transparency and reproducibility for grammar-engineering experiments. More importantly, Lam and Uí Dhonnchadha (2026) found that around 70% of its generated Cantonese LFG f-structures captured key predicate–argument relations, suggesting some structural competence relevant to grammar engineering. We compared its performance with `GPT-5.4`, OpenAI's frontier proprietary model at the time of this study. This comparison also allows us to examine potential differences between open-weight and proprietary frontier LLMs in formal grammar generation, as previous studies have generally found proprietary models to outperform open-weight ones (e.g., Devine et al., 2026). We set the temperature to 0 and kept `top_p` as 1.0 to minimize stochastic variation and encourage reproducible generation.

**Experimental design and prompts** We designed two experiments to evaluate whether LLMs can assist in the creation of Cantonese XLE grammars under different prompting conditions. The experiments varied along two dimensions: (i) the source of grammar formation (sentences vs. target f-structures), and (ii) the scope of grammatical coverage targeted by the prompt (single vs. multiple construction settings), as illustrated in Table 1. Both experiments used the Cantonese ParGram resources (Section 3) as gold standards. The models were asked to generate XLE-style grammars containing phrase structure rules, functional annotations, templates where appropriate, and lexical entries. All prompts included a one-shot English toy grammar in XLE format as an in-context example of the expected grammar representation. The inclusion of the English toy grammar simulates the ParGram setting in which grammar engineers are first introduced to grammar engineering through English toy grammars before developing grammars for their own languages. Full prompts are provided in Appendices C–F.

Experiment 1 evaluated single-construction grammar creation. For each of the 50 Cantonese ParGram sentences (Appendix A), we compared two prompting conditions: sentence-to-grammar and f-structure-to-grammar. In the former, the model was given only the Cantonese sentence; in the latter, the corresponding target f-structure from the Cantonese ParGram treebank was provided. This experiment tested whether access to the target syntactic representation improved the quality of the generated grammar. Both conditions were evaluated with `gpt-oss-120b` and `GPT-5.4`, yielding a $2 \times 2$ design crossing model type with source of grammar formation. English baselines using `GPT-5.4` on the corresponding English ParGram sentences (Butt et al., 1999; Rosén et al., 2012; Sulger et al., 2013) were included to examine whether similar effects could also be observed for a higher-resource language (English).[3]

| Experiment | Prompt Condition | Models | Grammar Count | Rule Count |
|---|---|---|---|---|
| Exp. 1 | Sentence → XLE Grammar | gpt-oss-120b, GPT-5.4 | 100 | 1602 |
| Exp. 1 | F-structure → XLE Grammar | gpt-oss-120b, GPT-5.4 | 100 | 1652 |
| Exp. 2 | Multiple Sentences → XLE Grammar | gpt-oss-120b, GPT-5.4 | 20 | 699 |
| Exp. 2 | Multiple F-structures → XLE Grammar | gpt-oss-120b, GPT-5.4 | 20 | 701 |

Table 1: Overview of Cantonese experiments. Grammar Count = total number of LLM-generated XLE grammars; Rule Count = total number of rules in LLM-generated XLE grammars.

Experiment 2 evaluated whether LLMs could generate more generalizable XLE grammars when provided with multiple constructions rather than a single target construction. This setting is more challenging because the generated grammar must simultaneously account for multiple sentences and complex interactions among grammatical constraints, as required in large-scale grammar engineering. The stimuli consisted of ten sets of Cantonese sentences, each combining a basic, an intermediate, and an advanced sentence drawn from the first 50 ParGram sentences.[4] As in Experiment 1, we compared two prompting conditions: multiple sentences-to-grammar and multiple f-structures-to-grammar. In the latter condition, the corresponding target f-structures were provided. This experiment, therefore, tested whether LLMs could move beyond sentence-specific analyses and generate XLE grammars to account for broader grammatical evidence. The same two models, `gpt-oss-120b` and `GPT-5.4`, were evaluated, again yielding a $2 \times 2$ design. An English `GPT-5.4` baseline was included for comparison.

**Evaluative criteria** Two types of evaluative ratings were used in the expert linguistic evaluation. ***Accuracy by Rule (AR)*** measures the percentage of accurate rules in each LLM-generated XLE grammar. A rule was considered accurate if it adhered to the LFG formalism and XLE notation requirements (Crouch et al., 2011), as exemplified in the prompts, including placement in the appropriate sections of the grammar.[5] However, formally correct rules do not necessarily guarantee successful parsing of the target construction(s), since grammars may contain rules that are formally correct but irrelevant to the required analysis. We therefore employed a ***Quality Rating (QR)*** scheme to assess the quality of the f-structures produced by the LLM-generated grammars. This rating scheme was adopted from Lam and Uí Dhonnchadha (2026) and consists of a four-point scale for evaluating LLM-generated f-structures: *Excellent* (fully correct), *Good* (structurally correct but containing incorrect features), *Fair* (partially correct analysis), and *Bad* (unrecognizable construction). The full description of each point is in Appendix G. In Section 4.2, the evaluation categories were coded ordinally (*Excellent* = 4, *Good* = 3, *Fair* = 2, *Bad* = 1) for statistical analysis.

### 4.2 Results and Discussion

Tables 2–4 summarize the main results for the experiments, while Table 5 reports the Wilcoxon signed-rank tests for statistical comparisons. Over-

[3]These English baseline conditions are not based on the toy English grammar included in the prompts. The experimental English items are the corresponding ParGramBank sentences and treebank f-structures generated using the existing large-scale English ParGram resources (Sulger et al., 2013); the toy grammar serves only as a one-shot illustration of XLE notation and formatting. Because the English and Cantonese items are parallel versions of the same ParGram test suite, the two sets of experimental materials are directly comparable.

[4]The three levels of sentential complexity were approximated using the number of grammatical functions (GFs) represented in the ParGram f-structures: 1–2 GFs = basic; 3–4 GFs = intermediate; $\geq$5 GFs = advanced.

[5]AR is a generated-rule accuracy measure rather than a precision/recall comparison against a fixed inventory of gold rules. Different well-formed XLE grammars may realize the same analysis through different decompositions of phrase-structure rules, templates, and lexical specifications; consequently, one-to-one alignment between generated and gold rules is not always well defined. For this reason, a classical rule-level precision/recall F1 score is not reported.

| Condition | Quality Rating (QR) | Accuracy by Rule (AR) | English Baseline QR | English Baseline AR |
|---|---|---|---|---|
| gpt-oss-120b Sentence-to-Grammar | 1.70 (1.00) | 72% | – | – |
| gpt-oss-120b F-structure-to-Grammar | 2.02 (2.00) | 75% | – | – |
| GPT-5.4 Sentence-to-Grammar | 2.26 (2.00) | 84% | 2.54 (2.00) | 91% |
| GPT-5.4 F-structure-to-Grammar | 2.88 (3.00) | 90% | 2.88 (3.00) | 92% |

Table 2: Results for Experiment 1 (single-construction grammar generation). QR reported as mean (median).

| Condition | Quality Rating (QR) | Accuracy by Rule (AR) | English Baseline QR | English Baseline AR |
|---|---|---|---|---|
| gpt-oss-120b Multi-Sent.-to-Gram. | 1.10 (1.00) | 73% | – | – |
| gpt-oss-120b Multi-F-str.-to-Gram. | 1.90 (2.00) | 76% | – | – |
| GPT-5.4 Multi-Sentence-to-Gram. | 2.00 (2.00) | 86% | 2.40 (2.00) | 91% |
| GPT-5.4 Multi-F-structure-to-Gram. | 2.50 (3.00) | 88% | 2.70 (3.00) | 94% |

Table 3: Results for Experiment 2 (multi-construction grammar generation). QR reported as mean (median).

| Condition | Model | Excellent | Good | Fair | Bad |
|---|---|---|---|---|---|
| **Experiment 1** | | | | | |
| Sentence-to-Grammar | gpt-oss-120b | 4% | 14% | 30% | 52% |
| Sentence-to-Grammar | GPT-5.4 | 20% | 20% | 26% | 34% |
| F-structure-to-Grammar | gpt-oss-120b | 4% | 14% | 62% | 20% |
| F-structure-to-Grammar | GPT-5.4 | 30% | 36% | 26% | 8% |
| **Experiment 2** | | | | | |
| Multi-Sentence-to-Gram. | gpt-oss-120b | 0% | 0% | 10% | 90% |
| Multi-Sentence-to-Gram. | GPT-5.4 | 0% | 10% | 80% | 10% |
| Multi-F-str.-to-Gram. | gpt-oss-120b | 0% | 10% | 70% | 20% |
| Multi-F-str.-to-Gram. | GPT-5.4 | 0% | 60% | 30% | 10% |

Table 4: Distribution of Cantonese Quality Ratings (QRs) across experimental conditions.

all, two patterns emerge. First, `GPT-5.4` outperformed `gpt-oss-120b` across both experiments, especially in QR. Second, prompts containing target f-structures generally led to better grammars than prompts containing only sentences.

**Experiment 1: single-construction grammar generation** In Experiment 1, `GPT-5.4` achieved higher scores than `gpt-oss-120b` in both prompting conditions. In the sentence-to-grammar condition, `GPT-5.4` reached a QR of 2.26 and an Accuracy by Rule (AR) of 84%, compared with 1.70 and 72% for `gpt-oss-120b`. The difference was significant for both QR and AR, with large effects (QR: $Z = 3.38$, adjusted $p = .002$, $r = .60$; AR: $Z = 4.27$, adjusted $p < .001$, $r = .62$). The advantage of `GPT-5.4` was even stronger in the f-structure-to-grammar condition, where it achieved a QR of 2.88 and an AR of 90%, compared with 2.02 and 75% for `gpt-oss-120b`. These model effects were again significant and large (QR: $Z = 5.31$, adjusted $p < .001$, $r = .88$; AR: $Z = 5.95$, adjusted $p < .001$, $r = .84$).

The distribution of QRs in Table 4 shows that the advantage of `GPT-5.4` was not merely a small increase in average score. In the sentence-to-grammar condition, 52% of `gpt-oss-120b` outputs were rated *Bad*, compared with 34% for `GPT-5.4`. In the f-structure-to-grammar condition, only 8% of `GPT-5.4` outputs were rated *Bad*, while 66% were rated either *Good* or *Excellent*. This suggests that the stronger model was more able to produce grammars that supported recognizable and structurally meaningful analyses.

Providing target f-structures improved performance in Experiment 1. For `gpt-oss-120b`, QR improved significantly from 1.70 to 2.02 ($Z = 2.66$, adjusted $p < .01$, $r = .49$), although the improvement in AR was not significant (72% to 75%; adjusted $p = .09$). For `GPT-5.4`, f-structure input improved both QR and AR, from 2.26 to 2.88 in QR and from 84% to 90% in AR (QR: $Z = 3.46$, adjusted $p = .002$, $r = .62$; AR: $Z = 3.22$, adjusted $p = .003$, $r = .48$).

The English baselines showed similar patterns in QR. Because the English conditions use corresponding items from established English ParGram resources, this parallel pattern suggests that the benefit of target f-structure input is not specific to Cantonese, although evaluation of more ParGram languages might be required for broader cross-linguistic generalization.

The findings suggest that target f-structures are useful not only as intended output representations but also as guidance for constructing grammars. The gap between QR and AR is important: a grammar may contain formally acceptable rules but still fail to produce the correct f-structure. This confirms the need to evaluate both rule-level accuracy and parse-level output quality.

**Experiment 2: multi-construction grammar generation** Experiment 2 tested a more difficult setting where the generated grammar had to cover multiple constructions simultaneously. Scores were generally lower than in Experiment 1, especially for `gpt-oss-120b`. In the multi-sentence-to-grammar condition, gpt-oss-120b achieved a QR of only 1.10, with 90% of outputs rated *Bad*. By contrast, GPT-5.4 achieved a QR of 2.00 and an AR of 86%, with most outputs rated *Fair*. The model effect was significant for both QR and AR (QR: $Z = 2.64$, adjusted $p < .05$, $r = .93$; AR: $Z = 2.40$, adjusted $p < .05$, $r = .76$), despite the smaller sample size of Experiment 2.

| Effect | Comparison | Metric | $Z$ | $p$ | Adjusted $p$ | Effect Size ($r$) |
|---|---|---|---|---|---|---|
| **Experiment 1 (Cantonese)** | | | | | | |
| Model | gpt-oss-120b vs. GPT-5.4 (Sentence-to-Grammar) | QR | 3.38 | $< .001$ | **0.002**** | 0.60 |
| Model | gpt-oss-120b vs. GPT-5.4 (Sentence-to-Grammar) | AR | 4.27 | $< .001$ | **< .001***** | 0.62 |
| Model | gpt-oss-120b vs. GPT-5.4 (F-structure-to-Grammar) | QR | 5.31 | $< .001$ | **< .001***** | 0.88 |
| Model | gpt-oss-120b vs. GPT-5.4 (F-structure-to-Grammar) | AR | 5.95 | $< .001$ | **< .001***** | 0.84 |
| Prompt | Sentence-to-Grammar vs. F-structure-to-Grammar (gpt-oss-120b) | QR | 2.66 | $< .001$ | **< .01**** | 0.49 |
| Prompt | Sentence-to-Grammar vs. F-structure-to-Grammar (gpt-oss-120b) | AR | 1.72 | 0.09 | 0.09 | 0.25 |
| Prompt | Sentence-to-Grammar vs. F-structure-to-Grammar (GPT-5.4) | QR | 3.46 | $< .001$ | **0.002**** | 0.62 |
| Prompt | Sentence-to-Grammar vs. F-structure-to-Grammar (GPT-5.4) | AR | 3.22 | $< .001$ | **0.003**** | 0.48 |
| Prompt | **English Baseline**: Sentence-to-Grammar vs. F-structure-to-Grammar (GPT-5.4) | QR | 2.11 | $< .05$ | **< .05*** | 0.48 |
| Prompt | **English Baseline**: Sentence-to-Grammar vs. F-structure-to-Grammar (GPT-5.4) | AR | 0.25 | 0.80 | 0.80 | 0.04 |
| **Experiment 2 (Cantonese)** | | | | | | |
| Model | gpt-oss-120b vs. GPT-5.4 (Multi-Sentence-to-Grammar) | QR | 2.64 | $< .01$ | **< .05*** | 0.93 |
| Model | gpt-oss-120b vs. GPT-5.4 (Multi-Sentence-to-Grammar) | AR | 2.40 | 0.02 | **< .05*** | 0.76 |
| Model | gpt-oss-120b vs. GPT-5.4 (Multi-F-structure-to-Grammar) | QR | 1.81 | 0.07 | 0.14 | 0.68 |
| Model | gpt-oss-120b vs. GPT-5.4 (Multi-F-structure-to-Grammar) | AR | 2.60 | $< .01$ | **< .05*** | 0.82 |
| Prompt | Multi-Sent.-to-Grammar vs. Multi-F-str.-to-Grammar (gpt-oss-120b) | QR | 2.43 | 0.01 | **< .05*** | 0.92 |
| Prompt | Multi-Sent.-to-Grammar vs. Multi-F-str.-to-Grammar (gpt-oss-120b) | AR | 0.87 | 0.38 | 0.52 | 0.31 |
| Prompt | Multi-Sentence-to-Grammar vs. Multi-F-structure-to-Grammar (GPT-5.4) | QR | 1.56 | 0.12 | 0.14 | 0.64 |
| Prompt | Multi-Sentence-to-Grammar vs. Multi-F-structure-to-Grammar (GPT-5.4) | AR | 1.13 | 0.26 | 0.52 | 0.38 |
| Prompt | **English Baseline**: Multi-Sent.-to-Grammar vs. Multi-F-str.-to-Grammar (GPT-5.4) | AR | 1.63 | 0.10 | 0.10 | 0.52 |

Table 5: Statistical comparisons for Experiments 1–2 using Wilcoxon signed-rank tests. QR = Quality Rating; AR = Accuracy by Rule. Adjusted $p$-values: Holm correction. $*p < .05$, $**p < .01$, $***p < .001$; **boldface** indicates statistical significance after correction ($p < .05$). English baselines: only 1 comparison per metric, adjusted $p$-value = raw $p$-value. QR for the English multi-construction baseline could not be computed because of insufficient non-zero pairs.

| Common Error Type | % Grammars | % Errors |
|---|---|---|
| **Annotated phrase-structure rule** | | |
| Incorrect GF in functional annotation | 22% | 7% |
| Incorrect part-of-speech | 20% | 7% |
| Duplicate rule | 24% | 8% |
| Unconstrained unary recursion | 8% | 3% |
| Missing functional annotation (or annotation conflict) | 28% | 9% |
| **Template** | | |
| Incorrect grammatical function (GF) | 28% | 9% |
| Missing template | 20% | 7% |
| Template formatting issue | 22% | 7% |
| **Lexical entry** | | |
| Incorrect PRED value | 16% | 5% |
| Missing / incorrect subcat. frame (GF) | 42% | 14% |
| Missing lexical entry rule | 10% | 3% |
| Misuse of template | 16% | 5% |
| Incorrect part-of-speech (or tokenization) | 8% | 3% |
| **Others**: Info specified at wrong location | 38% | 13% |

Table 6: Error analysis for Experiment 1 (gpt-oss-120b, Cantonese, Sentence-to-Grammar). % Grammars is relative to all generated grammars ($N = 50$). Because a single grammar may contain multiple error types, percentages in this column do not sum to 100%. % Errors is relative to the total number of identified error type instances ($N = 151$) spread across the generated grammars and therefore sums to 100%.

In the multi-f-structure-to-grammar condition, both models improved, but in different ways. gpt-oss-120b improved from 1.10 to 1.90 in QR, and the prompt-source effect was significant for QR ($Z = 2.43$, adjusted $p < .05$, $r = .92$), though not for AR. GPT-5.4 reached higher scores, with a QR of 2.50 and an AR of 88%; 60% of its outputs were rated *Good*. However, the prompt-source effects for GPT-5.4 were not significant. Given the relatively small sample size in Experiment 2, this may partly reflect limited statistical power rather than the complete absence of a prompt-source effect, especially since the corresponding effect sizes remained moderate (AR: $r = .38$) to large (QR: $r = .64$). This suggests that while GPT-5.4 was able to use surface-sentence evidence more effectively than gpt-oss-120b, target f-structures still improved the overall descriptive pattern. The English baselines pointed in the same direction.

**Common error analysis** Table 6 presents a detailed error analysis for the lowest-performing Cantonese condition, namely gpt-oss-120b in Experiment 1 sentence-to-grammar. The most frequent errors involved missing or incorrect subcategorization frames, information specified in the wrong section of the grammar, missing functional annotations, incorrect GFs, and missing templates. Similar error types were observed in other conditions, although generally less frequently. Overall, the results suggest that the main difficulty was not generating XLE-like notation, but coordinating interactions among phrase-structure rules, lexical entries, templates, and functional constraints for linguistic analysis. In relation to Section 3's con-structions, for functional control, `GPT-5.4` could generate the required structure-sharing equations, although they were occasionally omitted. For 將 object-fronting, both models produced some OBL analyses, but neither effectively employed lexical rules to capture the alternation. For passivization, both models could generate a three-place subcategorization frame for 俾 when prompted with the target f-structure, but neither successfully formulated the structure-sharing equations.

**Implications for grammar engineering and human–AI collaboration** Lam and Uí Dhonnchadha (2026) showed that `gpt-oss-120b` could generate Cantonese LFG f-structures capturing key predicate–argument relations in around 70% of cases. The present study further indicates that generating grammars directly from sentences remains substantially more difficult, whereas prompts containing target f-structures produced better grammars. Taken together, these findings point to a possible workflow where LLMs first generate preliminary f-structures from sentence inputs, after which engineers verify the analyses and use the corrected f-structures to prompt grammar generation. The resulting grammars can undergo refinement.

Such a workflow places human linguistic expertise at the center rather than treating LLMs as replacements for grammar engineers. The models may assist with intermediate analyses and the generation of candidate grammar components, while experts contribute the theoretical analysis, formal validation, and cross-constructional reasoning required for robust grammar development. In this sense, the potential of LLMs for grammar engineering lies particularly in supporting expert grammar engineers rather than in fully autonomous grammar development.

## 5 Conclusion

This paper presented new Cantonese ParGram resources and a controlled experimental evaluation of LLM capabilities in supporting knowledge-driven grammar engineering, with corresponding English baselines. `GPT-5.4` outperformed `gpt-oss-120b`; prompts containing target f-structures generally produced better grammars than prompts containing sentences. Both models struggled to coordinate interactions among phrase-structure rules, lexical entries, templates, and functional constraints, especially in multi-construction settings. The findings characterize both the capabilities and limitations of current LLMs for grammar engineering: LLMs may provide support under expert supervision, while human linguistic expertise remains central to robust grammar development.

## Limitations

Several limitations should be noted. First, the experiments focused on only two OpenAI models, namely `gpt-oss-120b` and `GPT-5.4`. The findings may therefore not generalize to other open-weight or proprietary LLMs. Second, although the Cantonese ParGram resources cover a broader range of grammatical phenomena, the experiments were limited to the first 50 ParGram sentences and a relatively small number of multi-construction sets. The statistical power of Experiment 2 was therefore limited by the small sample size. Third, the study focused exclusively on in-context learning through one-shot prompting and did not investigate fine-tuning approaches, which may improve grammar generation performance. Fourth, the evaluation focused mainly on the formal well-formedness of the generated XLE grammars, rather than computational efficiency or broader downstream NLP performance. Finally, although Experiment 2 extends the controlled experimental paradigm to multi-construction grammar generation, the experimental grammars remain substantially smaller than mature broad-coverage ParGram grammars, such as those developed for English, German, and Polish, which involve substantially larger inventories of interacting rules and constraints. As in controlled experimental paradigms more generally, the purpose of the present experiments is not to reproduce the full complexity of the target system, but to isolate specific variables in a tractable setting and systematically evaluate their effects on LLM grammar generation. The controlled paradigm therefore provides an empirical basis for identifying capabilities and limitations that can subsequently be investigated in increasingly large-scale grammar-development settings.

## Acknowledgments

I gratefully acknowledge the support of a Research Project Grant from the Leverhulme Trust for the project *Modelling Coreference Resolution across Syntax, Semantics, and Discourse* (RPG-2025-064). I am very grateful to Mary Dalrymple and Elaine Uí Dhonnchadha for their feedback during the preparation of this paper. I would also like to thank members of the ParGram f-structure comparison meetings for their continued feedback and support since 2023. Finally, I greatly appreciate the three anonymous reviewers and the Area Chair for their helpful feedback, as well as the Program Committee for their consideration of this work.

## A ParGramBank Sentences

**English ParGram Sentences (1–50):**
These sentences are openly accessible at: https://clarino.uib.no/iness/home > Treebanks > English and ParGram > View and search the selected treebanks.

1. The driver starts the tractor.
2. The tractor is red.
3. What did the farmer see?
4. Did the farmer sell his tractor?
5. Push the button.
6. Don't push the button.
7. The farmer gave his neighbor an old tractor.
8. The farmer cut the tree down.
9. The farmer groaned.
10. My neighbor was given an old tractor by the farmer.
11. The tree was cut down yesterday.
12. The tree had been cut down.
13. The tractor starts with a shudder.
14. The tractor appeared.
15. The boy knows the tractor is red.
16. The child thinks he started the tractor.
17. The farmer knows who started the tractor.
18. The child wondered whether the button had been pushed.
19. The tractor that the farmer bought is red.
20. The man who bought the tractor left.
21. The store the farmer bought the tractor from closed.
22. Whoever bought this tractor is a lucky person.
23. The farmer made his son clean the tractor.
24. The farmer made his son leave.

25. The farmer made her son buy the tractor.
26. The farmer let her son buy the tractor.
27. The farmer bought his son a tractor.
28. The woman bought the tractor for her husband.
29. The lovers danced until dawn.
30. The boy bathed in the river.
31. The teacher read to himself aloud.
32. The brothers bought the tractor for each other.
33. My sister is a great teacher.
34. The child is in the house.
35. The children are happy.
36. It is raining.
37. There is a problem with the tractor.
38. The book cover depicted a tractor.
39. Let's get ice-cream.
40. The boy swept up the broken wine bottle.
41. The red wine bottle broke.
42. There are great green globs of greasy grimy gopher guts.
43. They are proud of their daughter.
44. My tractor is faster than your sports car.
45. My tractor is the fastest vehicle in the county.
46. The barking dog woke the neighbours.
47. The tea drinking woman admired her new purchase from eBay.
48. The farmer wants to buy a tractor.
49. The farmer's daughter promised to repair the tractor.
50. The farmer persuaded his wife to buy a new tractor.

**Cantonese ParGram Sentences (1–50):**
The following sentences are the idiomatic Cantonese translation of the respective English ParGram sentences.

1. 個司機啓動拖拉機
2. 架拖拉機係紅色嘅
3. 個農夫睇到乜嘢
4. 個農夫賣咗佢嘅拖拉機呀
5. 撳個制
6. 唔好撳個制
7. 個農夫送畀佢嘅隔離屋一架舊拖拉機
8. 個農夫將棵樹斬落嚟
9. 個農夫嘆氣
10. 個農夫送咗一架舊拖拉機畀我隔離屋
11. 棵樹琴日俾人斬咗落嚟
12. 棵樹俾人斬咗落嚟
13. 個拖拉機顫咗一陣之後啓動咗
14. 架拖拉機出現咗
15. 個男仔知道架拖拉機係紅色嘅
16. 個細路仔以為佢啓動咗架拖拉機
17. 個農夫知道邊個啓動咗架拖拉機
18. 個細路仔想知個制係咪俾人撳咗落去
19. 個農夫買嘅拖拉機係紅色嘅
20. 買嗰架拖拉機嘅男人走咗
21. 個農夫買拖拉機嘅舖頭收咗檔
22. 買咗呢架拖拉機嘅人好幸運
23. 個農夫叫佢個仔清理拖拉機
24. 個農夫叫佢個仔走
25. 個農夫叫佢個仔買咗一架拖拉機
26. 個農夫俾佢個仔買咗一架拖拉機
27. 個農夫幫佢個仔買咗一架拖拉機
28. 個女人買咗呢架拖拉機畀佢老公
29. 呢對情侶一直跳舞到天光
30. 個男仔喺條河度沖涼
31. 個老師大聲咁讀畀自己聽
32. 兄弟互相買咗架拖拉機

33. 我家姐係一位好老師

34. 嗰個細路喺屋入面

35. 啲細路好開心

36. 而家落緊雨

37. 架拖拉機有問題

38. 書嘅封面畫咗一架拖拉機

39. 我哋去食雪糕啦

40. 個男仔將酒樽嘅碎片掃起嚟

41. 個紅酒樽爆咗

42. 有一大團油淋淋又污糟嘅地鼠內臟

43. 佢哋為個女覺得自豪

44. 我架拖拉機比你架跑車快

45. 我架拖拉機係成個縣最快嘅交通工具

46. 隻吠緊嘅狗嘈醒咗啲鄰居

47. 個飲緊茶嘅女人欣賞佢喺 eBay 買返嚟嘅新嘢

48. 個農夫想買一架拖拉機

49. 個農夫嘅女應承修理架拖拉機

50. 個農夫勸佢老婆買一架新拖拉機

## B Mapping Algorithm: C-to-f Structure Mapping

The following shows the mapping from the c-structure to the f-structure of the sentence *The boys love cats* via LFG's mapping algorithm.

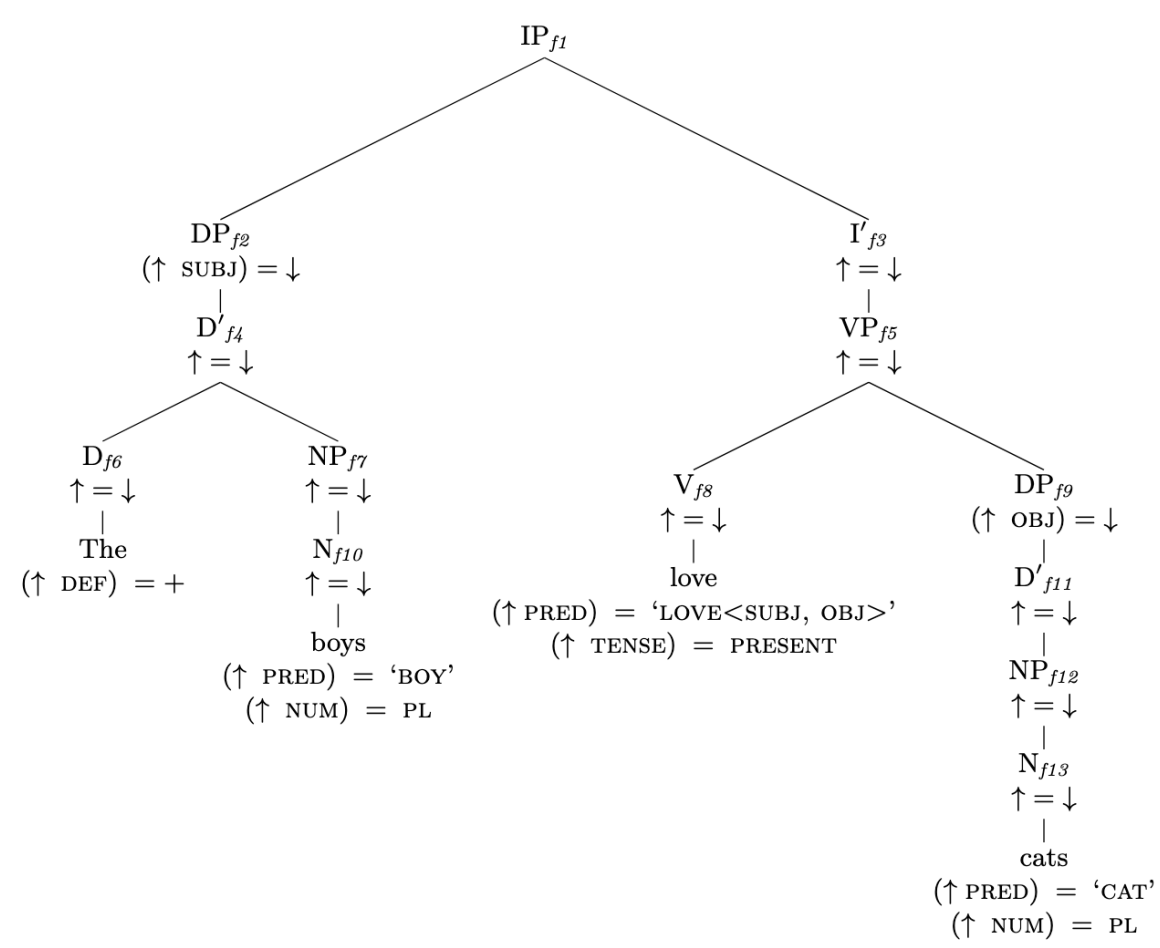

Figure 4: C-structure. Labels (e.g., $f_1$, $f_2$) are provided in the c-structure, where $f_1$ refers to the f-structure associated with the IP node and so on.

**Resolving Functional Equations**

$$(f_1\ \text{SUBJ}) = f_2 \qquad f_1 = f_3 \tag{1}$$

$$f_2 = f_4 \qquad f_4 = f_6 \qquad f_4 = f_7 \qquad f_7 = f_{10} \tag{2}$$

i.e., $f_2 = f_4 = f_6 = f_7 = f_{10}$

$$f_3 = f_5 \qquad f_5 = f_8 \qquad (f_5\ \text{OBJ}) = f_9 \qquad f_9 = f_{11} \qquad f_{11} = f_{12} \qquad f_{12} = f_{13} \tag{3}$$

i.e., $f_3 = f_5 = f_8$
i.e., $(f_5\ \text{OBJ}) = f_9 = f_{11} = f_{12} = f_{13}$

$$\begin{aligned} (f_6\ \text{DEF}) &= + \\ (f_{10}\ \text{PRED}) &= \text{`BOY'} \\ (f_{10}\ \text{NUM}) &= \text{PL} \\ (f_8\ \text{PRED}) &= \text{`LOVE<SUBJ, OBJ>'} \\ (f_8\ \text{TENSE}) &= \text{PRESENT} \\ (f_{13}\ \text{PRED}) &= \text{`CAT'} \\ (f_{13}\ \text{NUM}) &= \text{PL} \end{aligned} \tag{4}$$

$$
f_1, f_3, f_5, f_8 \begin{bmatrix} \text{PRED} & \text{`LOVE<SUBJ, OBJ>'} \\ \text{TENSE} & \text{PRESENT} \\ \text{SUBJ} & f_2, f_4, f_6, f_7, f_{10} \begin{bmatrix} \text{PRED} & \text{`BOY'} \\ \text{DEF} & + \\ \text{NUM} & \text{PL} \end{bmatrix} \\ \text{OBJ} & f_9, f_{11}, f_{12}, f_{13} \begin{bmatrix} \text{PRED} & \text{`CAT'} \\ \text{NUM} & \text{PL} \end{bmatrix} \end{bmatrix}
$$

Figure 5: F-structure. The f-structure is formed by resolving the functional equations annotated on the c-structure. Labels (e.g., $f_1$, $f_2$) are then omitted after c-to-f-structure mapping.

# C An Example Prompt: From Cantonese Sentence to XLE Grammar

All prompts consist of four parts: (i) system prompt; (ii) user prompt; (iii) English toy grammar; (iv) final user instruction. The prompts were designed to simulate the ParGram training setting in which grammar engineers are first introduced to XLE grammar engineering through English toy grammars before developing grammars for their own languages.

**Part I: System Prompt**

You are an expert in Cantonese, English, Lexical-Functional Grammar (LFG), and Xerox Linguistic Environment (XLE) for grammar engineering. You produce linguistically well-formed XLE grammars with accurate phrase structure rules, functional annotations, templates, and lexical entries. Follow standard XLE notation and formatting conventions. Your output must be concise, syntactically consistent, and suitable for parsing in XLE.

**Part II: User Prompt**

Your task is to create an XLE grammar for Cantonese. The grammar must be able to parse the Cantonese sentence enclosed in triple backticks.

Your grammar must contain:

1. Phrase structure rules annotated with functional information
2. A Cantonese lexicon with parts of speech and functional annotations

You may optionally include templates if appropriate.

Do not generate:

explanations
commentary
c-structures
f-structures

Only output the grammar itself.

**Part III: English Toy Grammar**

```
Below is an example of an English XLE grammar
that parses the sentence:

"Sam liked the apples."

DEMO ENGLISH RULES (1.0)

IP --> DP: (^SUBJ)=!;
I':^=!.

I' --> (I:^=!)
VP:^=!.

VP --> V:^=!;
DP:(^OBJ)=!.

DP --> (D:^=!);
NP:^=!.

NP --> N:^=!.

DEMO ENGLISH TEMPLATES (1.0)

TRANS(P)=@(PASS(^PRED)='P<(^SUBJ)(^OBJ)>').

PASS(FRAME)={FRAME
            (^PASSIVE)=-
            |FRAME
            (^PASSIVE)=+
            (PARTICIPLE)=c past
            (^OBJ)-->(^SUBJ)
            {(^SUBJ)-->(^OBL-AG)
            |(^SUBJ)-->NULL
            }}.

DEMO ENGLISH LEXICON (1.0)

Sam N * (^PRED)='sam'.

liked V * @(TRANS like)
"or (^PRED)='like<(^SUBJ)(^OBJ)>' without
template"
          (^TENSE)=past.

the D * (^DEF)=+.

apples N * (^PRED)='apple'
           (^NUM)=pl.
```

**Part IV: Final User Instruction**

Now create an XLE grammar for the following Cantonese sentence:

'''個司機啓動拖拉機'''

## D An Example Prompt: From Cantonese F-structure to XLE Grammar

All prompts consist of four parts: (i) system prompt; (ii) user prompt; (iii) English toy grammar; (iv) final user instruction. The prompts were designed to simulate the ParGram training setting in which grammar engineers are first introduced to XLE grammar engineering through English toy grammars before developing grammars for their own languages.

Please refer to Appendix C for the system prompt and English toy grammar. An example of the final user instruction for the Cantonese f-structure-to-grammar conditions is as follows:

**Part II: User Prompt**

Your task is to create an XLE grammar for Cantonese. The grammar must be able to parse a Cantonese sentence and produce the exact f-structure enclosed in triple backticks.

Your grammar must contain:

1. Phrase structure rules annotated with functional information
2. A Cantonese lexicon with parts of speech and functional annotations

You may optionally include templates if appropriate.

Do not generate:

explanations
commentary
c-structures
f-structures

Only output the grammar itself.

**Part IV: Final User Instruction**

Now create an XLE grammar capable of producing the following Cantonese f-structure:

```
'''
[ PRED   '啓動<SUBJ, OBJ>'
  TNS-ASP [ PROG  -
            PERF  -
            EXP   - ]
  OBJ    [ PRED   '拖拉機'
           NTYPE  [ NSYN   common ]
           PERS   3
           ANIM   - ]
  SUBJ   [ PRED   '司機'
           SPEC   [ CLASS-MEASURE
                      [ PRED   '個'
                  TYPE  classifier ] ]
           NTYPE  [ NSYN   common ]
           PERS   3
           HUMAN  + ]
  VTYPE  main ]
'''
```

## E An Example Prompt: From English Sentence to XLE Grammar

All prompts consist of four parts: (i) system prompt; (ii) user prompt; (iii) English toy grammar; (iv) final user instruction.

**Part I: System Prompt**

You are an expert in English, Lexical-Functional Grammar (LFG), and Xerox Linguistic Environment (XLE) for grammar engineering. You produce linguistically well-formed XLE grammars with accurate phrase structure rules, functional annotations, templates, and lexical entries. Follow standard XLE notation and formatting conventions. Your output must be concise, syntactically consistent, and suitable for parsing in XLE.

**Part II: User Prompt**

Your task is to create an XLE grammar for English. The grammar must be able to parse the English sentence enclosed in triple backticks.

Your grammar must contain:

1. Phrase structure rules annotated with functional information
2. An English lexicon with parts of speech and functional annotations

You may optionally include templates if appropriate.

Do not generate:

explanations
commentary
c-structures
f-structures

Only output the grammar itself.

**Part III: English Toy Grammar**

```
Below is an example of an English XLE grammar
that parses the sentence:

"Sam liked the apples."

DEMO ENGLISH RULES (1.0)

IP --> DP: (^SUBJ)=!;
I':^=!.

I' --> (I:^=!)
VP:^=!.

VP --> V:^=!;
DP:(^OBJ)=!.

DP --> (D:^=!);
NP:^=!.

NP --> N:^=!.

DEMO ENGLISH TEMPLATES (1.0)

TRANS(P)=@(PASS(^PRED)='P<(^SUBJ)(^OBJ)>').

PASS(FRAME)={FRAME
            (^PASSIVE)=-
            |FRAME
            (^PASSIVE)=+
            (PARTICIPLE)=c past
            (^OBJ)-->(^SUBJ)
            {(^SUBJ)-->(^OBL-AG)
            |(^SUBJ)-->NULL
            }}.

DEMO ENGLISH LEXICON (1.0)

Sam N * (^PRED)='sam'.

liked V * @(TRANS like)
"or (^PRED)='like<(^SUBJ)(^OBJ)>' without
template"
          (^TENSE)=past.

the D * (^DEF)=+.

apples N * (^PRED)='apple'
           (^NUM)=pl.
```

**Part IV: Final User Instruction**

Now create an XLE grammar for the following English sentence:

```
'''The driver starts the tractor'''
```

## F An Example Prompt: From English F-structure to XLE Grammar

All prompts consist of four parts: (i) system prompt; (ii) user prompt; (iii) English toy grammar; (iv) final user instruction.

Please refer to Appendix E for the system prompt and English toy grammar. An example of the final user instruction for the English f-structure-to-grammar condition is as follows:

**Part II: User Prompt**

Your task is to create an XLE grammar for English. The grammar must be able to parse an English sentence and produce the exact f-structure enclosed in triple backticks.

Your grammar must contain:

1. Phrase structure rules annotated with functional information
2. An English lexicon with parts of speech and functional annotations

You may optionally include templates if appropriate.

Do not generate:

explanations
commentary
c-structures
f-structures

Only output the grammar itself.

**Part IV: Final User Instruction**

Now create an XLE grammar capable of producing the following English f-structure:

```
'''
[ PRED        'start<SUBJ,OBJ>'
  TNS-ASP    [ TENSE  pres
               PROG   -
               PERF   -
               MOOD   indicative ]
  OBJ         [ PRED    'tractor'
                SPEC    [ DET
                           [ PRED      'the'
                            DET-TYPE  def ] ]
             NTYPE   [ NSEM   [ COMMON  count ]
                              NSYN   common ]
                PERS    3
                NUM     sg
                CASE    obl ]
  SUBJ        [ PRED    'driver'
                SPEC    [ DET
                           [ PRED      'the'
                            DET-TYPE  def ] ]
             NTYPE   [ NSEM   [ COMMON  count ]
                              NSYN   common ]
                PERS    3
                NUM     sg
                CASE    nom ]
  VTYPE       main
  PASSIVE     -
  CLAUSE-TYPE decl ]
'''
```

These English treebank f-structures are openly accessible at: https://clarino.uib.no/iness/home > Treebanks > English and ParGram > View and search the selected treebanks.

## G Quality Rating (QR): Four-point Scale

We employed a Quality Rating (QR) scheme to assess the quality of the f-structures produced by the LLM-generated grammars. This rating scheme was adopted from Lam and Uí Dhonnchadha (2026, p. 134) and consists of a four-point scale as follows for evaluating LLM-generated f-structures:

- Excellent (Fully correct)
  - All grammatical functions (GFs) are correctly specified (e.g. SUBJ, OBJ, OBL, COMP, XCOMP, ADJUNCT)
  - Subcategorization frame <...> is correct
  - All relevant non-GF features (e.g., TENSE, ASPECT, NUMBER, GENDER, CASE, PERSON) are correct

- Good (Structurally correct, some incorrect features)
  - All GFs are correct and consistent with the predicate argument structure
  - Subcategorization frame is correct
  - Only issues in non-GF features

- Fair (Partially correct analysis)
  - The core construction is recognizable, but there is at least one GF error, such as a missing argument or incorrect GF assignment (e.g. OBJ vs OBL) or a mismatch with the subcategorization frame
  - May include errors in non-GF features

- Bad (Not recognizable construction)
  - The core construction is not recognizable due to major errors in GFs
  - The f-structure is incompatible with the intended predicate-argument structure; e.g., relative clause or adjunct analyzed as complement